\documentclass[runningheads]{llncs}

\usepackage{graphicx}
\usepackage{multirow}
\usepackage{amsmath}
\usepackage{subcaption}
\usepackage{url}

\begin{document}
\title{Underestimation Bias and Underfitting \\ in Machine Learning}

%
%
\author{P\'{a}draig Cunningham\inst{1}
\and
Sarah Jane Delany\inst{2}
}
\authorrunning{Cunningham \& Delany}
%
\institute{University College Dublin \\
\email{padraig.cunningham@ucd.ie}
\and
Technological University Dublin \\
\email{sarahjane.delany@tudublin.ie}
}
\maketitle              
\begin{abstract}
Often, what is termed algorithmic bias in machine learning will be due to historic bias in the training data. But sometimes the bias may be introduced (or at least exacerbated) by the algorithm itself. The ways in which algorithms can actually accentuate bias has not received a lot of  attention with researchers focusing directly on methods to eliminate bias - no matter the source. In this paper we report on initial research to understand the factors that contribute to bias in classification algorithms. We believe this is important because underestimation bias is inextricably tied to regularization, i.e. measures to address overfitting can accentuate bias.  

\end{abstract}
\section{Introduction}

Research on bias in Machine Learning (ML) has focused on two issues; how to measure bias and how to ensure fairness \cite{menon2018cost}. In this paper we examine the contribution of the classifier algorithm to bias. 
An algorithm would be biased if it were more inclined to award desirable outcomes to one side of a sensitive category. The desirable outcome could be a loan or a job, the category might be gender or race. 
It is clear that there are two main sources of bias in classification \cite{kamishima2012fairness}:
\begin{itemize}
    
    \item \textbf{Negative Legacy}: the bias is there in the training data, either due to poor sampling, incorrect labeling or discriminatory practices in the past. 

    \item \textbf{Underestimation}: the classifier under-predicts an infrequent outcome for the minority group. This can occur when the classifier 
    \emph{underfits} the data, thereby focusing on strong signals in the data and missing more subtle phenomena. 
    \end{itemize}
In most cases the data (negative legacy) rather than the algorithm itself is the root of the problem. This question is neatly sidestepped in most fairness research by focusing on fair outcomes no matter what is the source of the problem. 

\begin{figure}[t]
    \centering
    \includegraphics[width=0.8\linewidth]{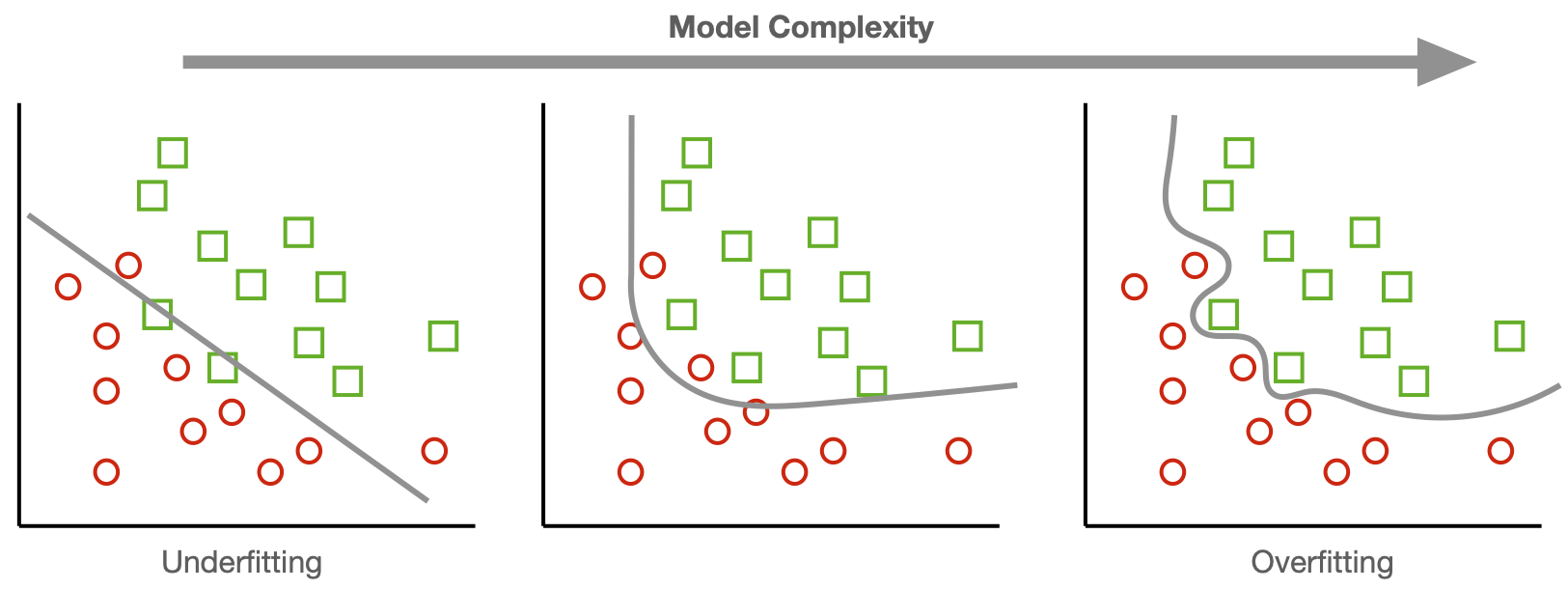}
  \caption{The relationship between model fitting and model complexity. The fitted model is shown as a grey line that separates two classes in the data (red circles and green squares). Overfitted models can fit to noise in the training data while underfitted models miss a lot of the detail.} \label{fig:underfit}
\end{figure}

We argue that it is useful to explore the extent to which algorithms accentuate bias because 
this issue is inextricably tied to regularisation, a central issue in ML. In developing ML models a key concern is to avoid overfitting. Overfitting occurs when the model fits to noise in the training data, thus reducing generalisation performance (see Figure \ref{fig:underfit}). Regularisation controls the complexity of the model in order to reduce the propensity to overfit. The way regularisation is achieved depends on the model. The complexity of decision trees can be controlled by limiting the number of nodes in the tree. Overfitting in neural networks can be managed by limiting the magnitude of the weights. In this paper we discuss models at the other end of the complexity spectrum that underfit the training data. We show that underestimation bias can occur with underfitted models. 

\section{Background}
Before proceeding to the evaluation it is worth providing some background on the key concepts discussed in this paper. In the next section we provide some definitions relating to regularisation. In section \ref{sec:bias} we review some relevant research on bias and fairness in ML. Section \ref{sec:IllCorr} introduces the concept of Illusory Correlation and discusses its relevance to algorithmic bias.

\subsection{Regularisation}
Regularisation is a term from Statistics that refers to policies and strategies to prevent overfitting. In training a supervised ML model, the objective it to achieve good generalisation performance on unseen data. This can be assessed by holding back some test data from the training process and using it to get an assessment of generalisation performance. The relative performance on these training and test sets provides some insight on  
 underfitting and overfitting \cite{Goodfellow-et-al-2016}:
\begin{itemize}
    \item \textbf{Underfitting} occurs when a model does not capture the detail of the phenomenon; these  models will have some errors on the training data (see Figure \ref{fig:underfit}).
    \item \textbf{Overfitting} occurs when the model fits too closely to the training data picking up detail that does not generalise to unseen data. An overfitted model will have low error on the training data but higher error on test data. 
\end{itemize}
\begin{figure}[t]
    \centering
    \includegraphics[width=0.5\linewidth]{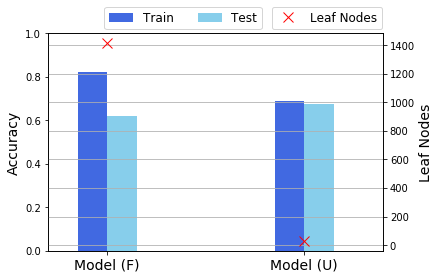}
  \caption{A demonstration of underfitting \textsf{Model(U)} and overfitting \textsf{Model(F)} on the Recidivism task presented in section \ref{sec:underest}.} \label{fig:fit}
\end{figure}

Controlling overfitting is one of the central concerns in ML. It is difficult because it depends on the interplay between the volume of data available and the \emph{capacity} of the model. If the training data provides comprehensive coverage of the phenomenon then overfitting is not a problem -- it need only be an issue when the training data is not comprehensive. The capacity of the model refers to its expressive power; low capacity models will underfit while high capacity models can overfit. Figure \ref{fig:fit} shows examples of under and overfitting on the Recidivism task described in section \ref{sec:underest}. In this case the model is a decision tree and the capacity is managed by controlling the number of leaf nodes. Limiting the number of leaf nodes to 30 results in an underfitted model whereas the overfitted model has been allowed to extend to 1413 leaf nodes. The overfitted model has good performance on the training data but does not generalise well. The underfitted model has better performance on the test (unseen) data.

\subsection{Bias and Fairness}\label{sec:bias}

The connection between regularisation and bias arises from the context in which bias occurs. Consider a scenario where the desirable classification outcome is the minority class (e.g. job offer) and the sensitive feature represents groups in the population where minority groups may have low \emph{base rates} for the desirable outcome \cite{barocas-hardt-narayanan}. So samples representing good outcomes for minority groups are scarce. Excessive regularisation causes the model to ignore or under-represent these data. 

We are not aware of any research on bias in ML that explores the relationship between underestimation and regularisation. This issue has received some attention (\cite{barocas-hardt-narayanan,kamishima2012fairness}) but it is not specifically explored. Instead research has addressed algorithmic interventions that ensure fairness as an outcome\cite{zemel2013learning,calders2010three,menon2018cost,zhang2019fairness}. 

Fundamental to all research on bias in ML are the measures to quantify bias. We define 
$Y$ to be the outcome (class variable) and $Y=1$ is the `desirable' outcome. 
$\hat{Y}= 1$ indicates that the classifier has predicted the desirable outcome. $S$
 is a `sensitive' feature and $S \ne 1$ is the minority group (e.g. non-white). In this notation the Calders Verwer \emph{discrimination score}\cite{calders2010three} is:
 \begin{equation}
\mathrm{CV} \leftarrow P[\hat{Y}= 1 | S = 1]-P[\hat{Y}= 1 | S \ne 1]
 \end{equation}
 If there is no discrimination, this score should be zero.

In contrast the \emph{disparate impact} ($\mathrm{DI}_S$) definition of unfairness \cite{feldman2015certifying} is a ratio rather than a difference:

\begin{equation}
   \mathrm{DI}_S \leftarrow \frac{P[\hat{Y}= 1 | S \ne 1]}{P[\hat{Y} = 1 \vert S = 1]} < \tau 
\end{equation}
$\tau = 0.8$ is the 80\% rule, i.e. outcomes for the minority should be within 80\% of those for the majority. 

For our evaluations we define an \emph{underestimation score} ($\mathrm{US}_{S}$) in line with $\mathrm{DI}_S$ that quantifies the underestimation effect described above:
\begin{equation}\label{eqn:US_S1}
    \mathrm{US}_{S} \leftarrow \frac{P[\hat{Y}= 1 | S \ne 1]}{P[Y = 1 | S \ne 1]} 
\end{equation}

If $\mathrm{US}_{S} <1$ the classifier is  predicting fewer desirable outcomes than are present in the training data. 

\subsection{Illusory Correlation}\label{sec:IllCorr}
Algorithmic bias due to underestimation is similar in some respects to the concept of Illusory Correlation in Psychology \cite{chapman1969illusory}. With Illusory Correlation people associate the frequent class with the majority group and the rare class with the minority, the infrequent class is overestimated for the minority group  \cite{costello2019rationality}. For instance, if the infrequent class is antisocial behaviour, the incidence will be over-associated with the minority. If the frequent class is a good credit rating, it will be over-associated with the majority. This over-estimation also happens for the frequent class in ML. However, the impact for the infrequent class and the minority feature is the opposite; the algorithm will likely accentuate the under-representation. In ML, the infrequent class outcome will be underestimated for the minority feature.

\begin{figure}
    \centering
    \includegraphics[width=0.8\linewidth]{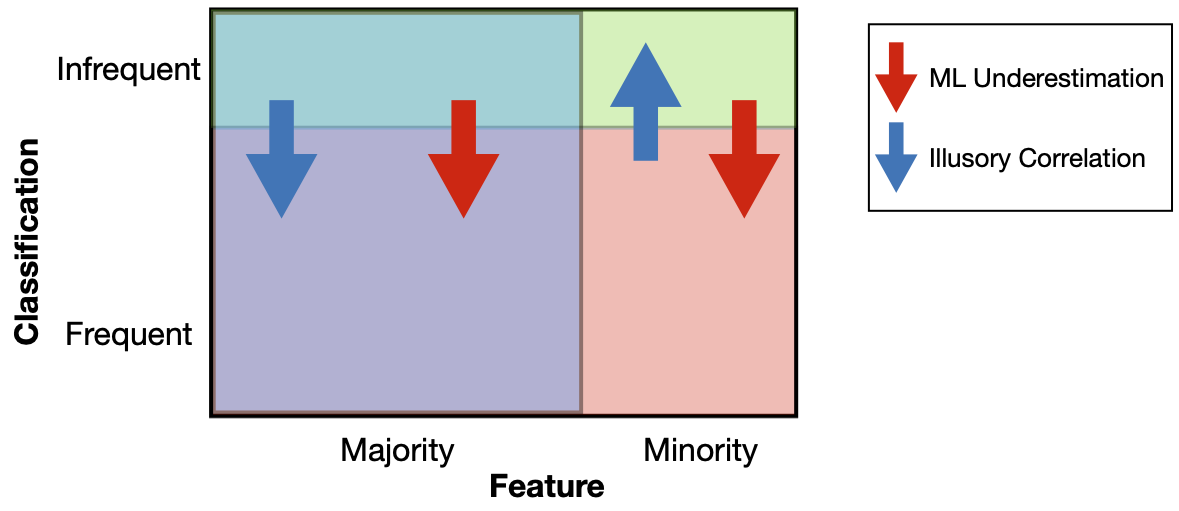}

  \caption{The relationship between Illusory Correlation and ML Underestimation. Illusory Correlation accentuates the frequent class for the majority and the infrequent class for the minority. With ML Underestimation the bias is all towards the Frequent outcome. 
  }\label{fig:IllCorr}
\end{figure}
In Figure \ref{fig:IllCorr} we see the similarities and differences between Illusory Correlation and ML bias. For the majority feature and the frequent class the behaviour is the same, the association is accentuated. However, the impact on the infrequent side of the classification is different, the ML bias is always \emph{towards} the frequent class. This is a problem if the infrequent outcome is the desirable one (e.g. loan approval, job offer). 
This difference is probably due to the different underlying mechanisms in humans and computers. Hamilton and Gifford  \cite{hamilton1976illusory} argue that Illusory Correlation is due to cognitive mechanisms that produce distortions in judgement and contribute to the formation of stereotypes. Whereas we argue here that underestimation results from a combination of poor data coverage and restricted model capacity. 

\section{Evaluation}\label{sec:eval}
We present preliminary results on three binary classification datasets from the UCI repository\footnote{\url{archive.ics.uci.edu/ml/}}. We also include an analysis on an anonymised and reduced version of the \emph{ProPublica} \textsf{Recidivism} dataset \cite{larson2016we,dressel2018accuracy}. Our analysis is based on the same set of seven features as used in the study by Dressel and Farid \cite{dressel2018accuracy}.
The UCI datasets are \textsf{Adult}, \textsf{Wholesale Customers} and \textsf{User Knowledge Modelling} (see Table \ref{tab:datasets}). 

We use the \textsf{Adult} and \textsf{Recidivism} datasets to demonstrate the relationship between underestimation and regularisation  (section \ref{sec:underest}) and then we explore this in more detail in the other two datasets. 
In this second part of the evaluation, in order  to control the propensity for bias, we introduce a sensitive binary feature \textsf{S} with \textsf{S $\ne$ 0} representing the sensitive minority. To show bias due to underestimation we set $P[S \ne 1|Y=1] = 0.15$ and $P[S \ne 1|Y\ne 1] = 0.3$, i.e. the sensitive group is under-represented in the desirable class. We also report baseline results with the incidence of \textsf{S $\ne$ 1} balanced with the sensitive group occurring 30\% of the time in both classes. 
\begin{table}
\begin{center}
\caption{The datasets used in the evaluation. For consistency, the Positive outcome is the minority class in all datasets. }\label{tab:datasets}
\begin{tabular}{ l | r | c| c| c}
\hline
Dataset & Samples & Features & \% Positive & Train : Test \\
\hline
Adult & 48,842 & 14 & 25\% & 2:1\\
Recidivism & 7,214 & 7& 45\% & 2:1 \\
Knowledge & 403 & 5 & 26\% & 1:2 \\
Wholesale & 440 & 7 & 32\% & 1:1 \\
\end{tabular}
\end{center}
\end{table}
We are interested in uncovering situations where this under-representation is accentuated by the classifier. In section \ref{sec:Algs} we look at the untuned performance of seven classifiers from \textsf{scikit-learn}. In section \ref{sec:underf} we look at the impact of regularisation on neural net performance. In the next section we provide a more formal account of our bias measures. 

\subsection{Evaluation Measures}\label{sec:evalmes}
If a binary classifier is biased this will show up as a mismatch between the predicted positives $\text{P}^\prime$ and the actual positives P (see Table \ref{tab:TPTN}). 
This is likely to happen when the training data is significantly imbalanced resulting in predictions that underestimate the overall minority class \cite{chawla2003smoteboost}. 
\begin{table}
\begin{center}
\caption{Confusion matrix for binary classification.}\label{tab:TPTN}
\begin{tabular}{ c c | c c | c}
& & \multicolumn{2}{c}{\textbf{Predicted}}\\
&  & Pos & Neg \\ 
  \hline
\multirow{2}{*}{\textbf{Actual}} & Pos & TP & FN & P \\  
                         & Neg & FP & TN & N  \\
\hline
&&$\text{P}^\prime$&$\text{N}^\prime$ 
\end{tabular}
\end{center}
\end{table}
From this perspective,  bias is independent of accuracy so whether predictions are correct or not (True or False) is not relevant. 
So the minority class bias is effectively an underestimation at a class level, i.e.:
\begin{equation}\label{eqn:Bias}
    \text{US} \leftarrow \frac{\text{TP+FP}}{\text{TP+FN}} = \frac{\text{P}^\prime}{\text{P}}
\end{equation}
When the classifier is biased away from the positive class, $\text{P}^\prime < \text{P}$ this US score is less than 1. The $\text{US}_{S}$ score defined in equation \ref{eqn:US_S1} measures the same fraction, but for samples where the sensitive feature $S \ne 1$.
\begin{equation}\label{eqn:US_S}
    \text{US}_{S} \leftarrow  \frac{\text{P}^\prime_{S \ne 1}}{\text{P}_{S \ne 1}}
\end{equation}
While bias can be considered independently of accuracy, there is an important interplay between bias and accuracy. So the final evaluation measure we consider is the overall accuracy: 
\begin{equation}\label{eqn:Acc}
    \text{Acc} \leftarrow \frac{\text{TP+TN}}{\text{TP+FN+FP+FN}}
\end{equation}


\subsection{Underestimation in Action}\label{sec:underest}

We use the \textsf{Adult} and \textsf{Recidivism} datasets to show the impact of underestimation. In Table \ref{tab:IllCorr} we see that in the \textsf{Adult} dataset 
Females with salaries greater than 50K account for just 4\% of cases and in the \textsf{Recidivism} dataset Caucasians are relatively underrepresented among repeat offenders. We will see that this under-representation is accentuated by the classifiers.

\begin{table}
 \caption{Summary statistics for the Adult and Recidivism datasets. In both cases a feature/class combination is significantly underrepresented in the data. Key feature-specific percentages are shown in red, for instance 11\% of females are in the $>50$K salary category.}\label{tab:IllCorr}
\centering
\includegraphics[width=0.85\linewidth]{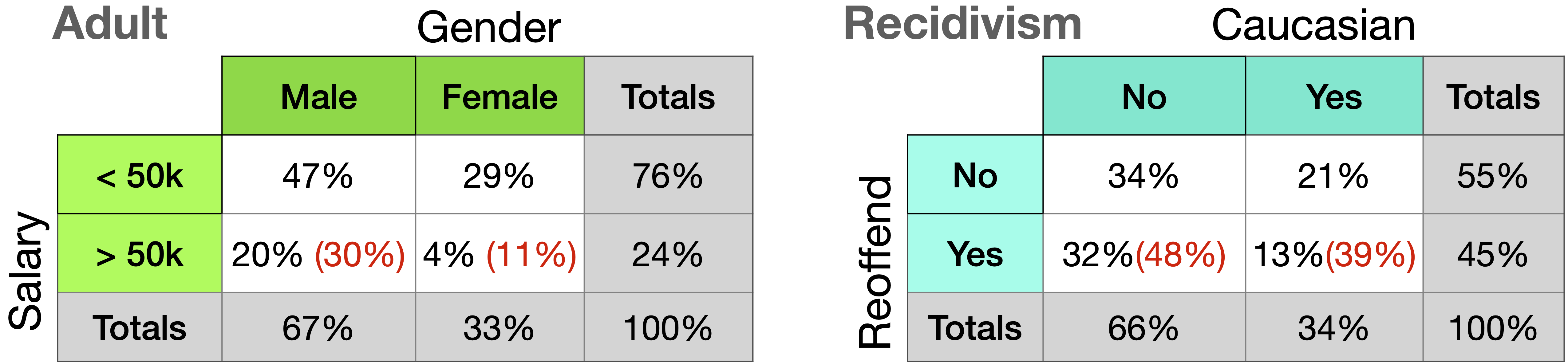}
\end{table}
\subsubsection{Adult Dataset:}This dataset is much studied in research on bias in ML because there is clear evidence of Negative Legacy \cite{calders2010three}. 
 At 33\%, females are underrepresented in the dataset. This under-representation is worse in the  $>50$K salary category where only $1/6$ are female.

\begin{figure}
  \begin{subfigure}[t]{.45\textwidth}
    \centering
    \includegraphics[width=\linewidth]{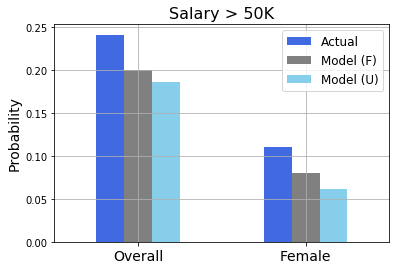}
    \caption{ }
  \end{subfigure}
  \hfill
  \begin{subfigure}[t]{.48\textwidth}
    \centering
    \includegraphics[width=\linewidth]{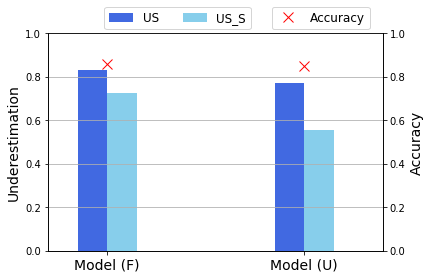}
    \caption{ }
  \end{subfigure}
  \caption{ A demonstration of model bias and underestimation on the \textsf{Adult} dataset. 
  (a) 24\% of all cases and 11\% of Females have Salary $>$ 50K. Both models underestimate these outcomes with the underfitted model \textsf{Model (U)} underestimating most. \\
  (b) Both models have similar accuracies and both underestimate overall (\textsf{US)}. The underestimation for Females (\textsf{US\_S}) is worse and is worst in the underfitted model \textsf{Model (U)}.
  }
\label{fig:adult}
\end{figure}

To illustrate underestimation we use a gradient boosting classifier \cite{prokhorenkova2018catboost}. We build two classifiers, one with just 5 trees (\textsf{Model (U)}) and one with 50 (\textsf{Model (F)}). Both models have good accuracy, 85\% and 86\% respectively.  Figure \ref{fig:adult}(a) shows the actual incidence of Salary $>$ 50 overall and for females. It also shows the predicted incidence by the two models. 
We can see that both models underestimate the probability of the Salary $>$ 50 class overall. On the right in Figure \ref{fig:adult}(a) we can see that this underestimation is exacerbated for females. 
This underestimation is worse in the underfitted model. The actual occurrence of salaries $>$ 50K for females is 11\% in the data, the underfitted model is predicting 6\%. The extent of this underestimation is quantified in Figure \ref{fig:adult}(b).

So underestimation is influenced by three things, underfitting, minority class and minority features. The underfitted model does a poor job of modelling the minority feature (female) in the minority class ($>50$K). This is not that easy to fix because it is often desirable not to allow ML models overfit the training data. 

\subsubsection{Recidivism Dataset:} We use the decision tree classifier from \textsf{scikit-learn}\footnote{\url{scikit-learn.org}} to demonstrate underestimation on the \textsf{Recidivism} dataset. In this case we control overfitting by constraining the size of the tree. The underfitted model (\textsf{Model (U)}) has 30 leaf nodes and the other model (\textsf{Model (F)}) has 1349 leaves. 
\begin{figure}[t]
  \begin{subfigure}[t]{.45\textwidth}
    \centering
    \includegraphics[width=\linewidth]{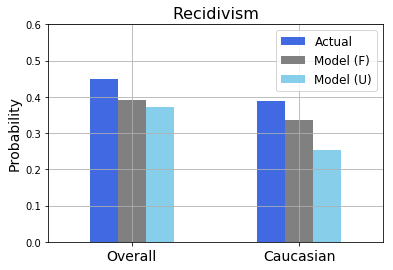}
    \caption{ }
  \end{subfigure}
  \hfill
  \begin{subfigure}[t]{.48\textwidth}
    \centering
    \includegraphics[width=\linewidth]{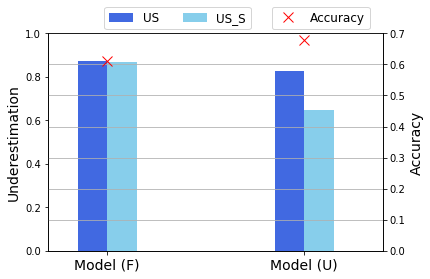}
    \caption{ }
  \end{subfigure}
  \caption{A demonstration of bias and underestimation on the \textsf{Recidivism} dataset. 
  (a) 46\% of all cases and 38\% of Caucasians are recidivists. Both models underestimate these outcomes with the underfitted model \textsf{Model (U)} underestimating most. \\
(b) The underfitted model has better generalisation accuracy. Both models underestimate overall (\textsf{US)} with the underfitted model underestimating more for Caucasians  (\textsf{US\_S}).}
\label{fig:recidivism}
\end{figure}
The picture here is similar but in this case the underfitted model has better accuracy. This accuracy comes at the price of increased underestimation. The underestimation is particularly bad for the minority feature with the level of recidivism for Caucasians significantly underestimated. As reported in other analyses  \cite{larson2016we,dressel2018accuracy} the input features do not provide a strong signal for predicting recidivism. So the fitted model does not generalise well to unseen data. On the other hand the model that is forced to underfit generalises better but fails to capture the Caucasian recidivism pattern. 

\subsection{Baseline Performance of Classifiers}\label{sec:Algs}
We move on now to look at the impact of a synthetic minority feature injected into the other two datasets. This synthetic feature is set up to be biased (negative legacy) $P[S \ne 1|Y=1] = 0.15$ and $P[S \ne 1|Y\ne 1] = 0.3$.
We test to see if this bias is accentuated (i.e. $US_S < 0$) for seven popular classifiers available in \textsf{scikit-learn}. For this assessment, the default parameters for the classifiers are used. There are two exceptions to this; the number of iterations for Logistic Regression and the Neural Network were increased to get rid of convergence warnings. 
\begin{figure}
  \begin{subfigure}[t]{.48\textwidth}
    \centering
    \includegraphics[width=\linewidth]{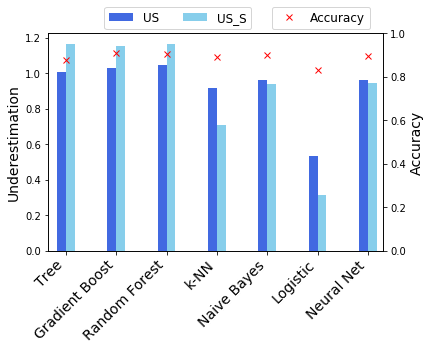}
    \caption{Wholesale dataset.}
  \end{subfigure}
  \hfill
  \begin{subfigure}[t]{.48\textwidth}
    \centering
    \includegraphics[width=\linewidth]{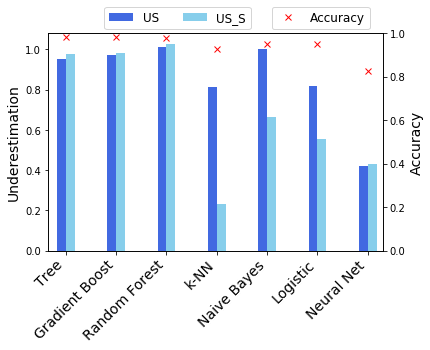}
    \caption{Knowledge dataset}
  \end{subfigure}
  \caption{The varying impact of underestimation across multiple classifiers. A sensitive feature S $\ne$ 1 has been added to both datasets (15\% in the desirable class and 30\% in the majority class. With the exception of the three tree-based classifiers, underestimation is exacerbated for this sensitive feature (\textsf{US\_S}). }\label{fig:mc}
\end{figure}
For the results  shown in Figure \ref{fig:mc} the main findings are:
\begin{itemize}
    \item The tree-based classifiers (Decision Tree, Gradient Boost \& Random Forest) all perform very well showing no bias (or a positive bias), both overall (US) and when we filter for the sensitive feature($\text{US}_S$). 
    \item The other classifiers ($k$-Nearest Neighbour, Naive Bayes, Logistic Regression \& Neural Network) all show bias (underestimation), overall and even more so in the context of the sensitive feature. 
    \item The accentuation effect is evident for the four classifiers that show bias, i.e. they predict even less than 15\% of the desirable class for the sensitive feature. 
\end{itemize}
This base-line performance by the tree based classifiers is very impressive. However, it is important to emphasise that the performance of the other methods can be improved significantly by parameter tuning -- as would be normal in configuring an ML system. Finally, it should not be inferred that tree-based methods are likely to be free of underestimation problems. In particular, the Decision Tree implementation in \textsf{scikit-learn} provides a number of mechanisms for regularisation that may introduce underestimation as a side effect. 

\begin{figure}[t]
  \begin{subfigure}[t]{.48\textwidth}
    \centering
    \includegraphics[width=\linewidth]{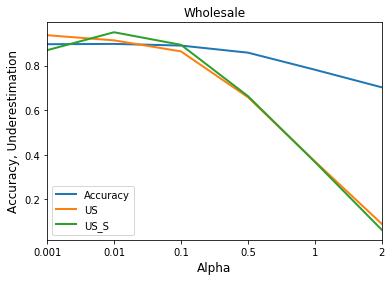}
    \caption{30\% in pos. class, 30\% in neg.}
  \end{subfigure}
  \hfill
  \begin{subfigure}[t]{.48\textwidth}
    \centering
    \includegraphics[width=\linewidth]{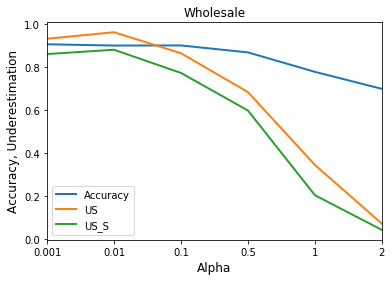}
    \caption{15\% in pos. class, 30\% in neg.}
  \end{subfigure}
   \medskip
  \begin{subfigure}[t]{.48\textwidth}
    \centering
    \includegraphics[width=\linewidth]{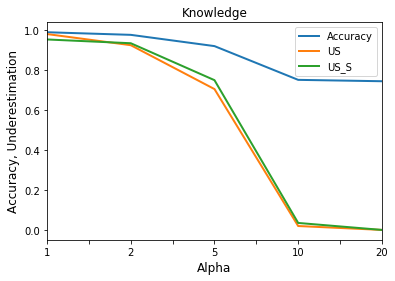}
    \caption{30\% in pos. class, 30\% in neg.}
  \end{subfigure}
  \hfill
  \begin{subfigure}[t]{.48\textwidth}
    \centering
    \includegraphics[width=\linewidth]{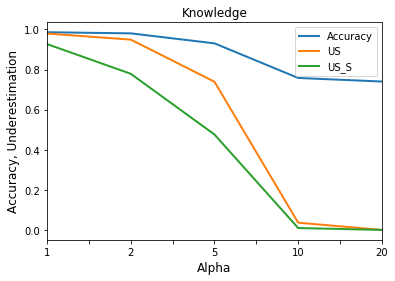}
    \caption{15\% in pos. class, 30\% in neg.}
  \end{subfigure}
\caption{The impact of underfitting on bias. A synthetic sensitive attribute has been added, it is underrepresented in the plots on the right (15\% compared with 30\%). 
Higher values of Alpha result in underfitting.  Furthermore, when the sensitive attribute is underrepresented in the desirable class (b) \& (d) the bias is exacerbated.}\label{fig:alpha}.
\end{figure}
\subsection{Impact of Underfitting}\label{sec:underf}
The standard practice in training a classifier is to ensure against overfitting in order to get good generalisation performance. Kamishima \emph{et al.} \cite{kamishima2012fairness} argue that bias due to underestimation arises when a classifier underfits the phenomenon being learned. This will happen when the data available is limited and samples covering the sensitive feature and the desirable class are scarce.

The \textsf{scikit-learn}\footnote{\url{scikit-learn.org}} neural network implementation provides an $\alpha$ parameter to control overfitting. It works by providing control over the size of the weights in the model. Constraining the weights reduces the potential for overfitting. The plots in Figure \ref{fig:alpha} show how underestimation varies with this $\alpha$ -- high values cause underfitting. These plots show Accuracy and Underestimation for different values of $\alpha$. For the plots on the left (Figure \ref{fig:alpha} (a)\&(c))
the incidence of the sensitive feature is the same in both the positive and negative class (30\%). For the plots on the right ((b)\&(d)) the sensitive feature is underrepresented (15\%) in the positive class. 

In Figure \ref{fig:alpha} (a) and (c) we see that high values of $\alpha$ (i.e. underfitting) result in significant bias. When the base rates for the minority group in the positive and negative classes are the same the US and $\text{US}_S$ rates are more or less the same.

When the prevalence of the sensitive group in the desirable class is lower ((b)\&(d)) the bias is exacerbated. It is important to emphasise that a good $\text{US}_S$ score simply means that underestimation is not present. There may still be bias against the minority group (i.e. poor CV or $\text{DI}_S$ scores). 

\section{Conclusions \& Future Work}
 In contrast to what Illusory Correlation tells us about how humans perceive things, underestimation occurs in ML classifiers because they are inclined to over-predict common phenomena and under-predict things that are rare in the training data. 
We have shown, on two illustrative datasets,  \textsf{Adult} and \textsf{Recidivism},  how the impact of under-representation in data leads to underestimation of the classifiers built on that data. 
We believe classifier bias due to underestimation is worthy of research because of its close interaction with regularisation.
We have demonstrated this interaction on two datasets where we vary the levels of under-representation and regularisation showing the impact on underestimation.  Underfitting data with an under-represented feature in the desirable class leads to increased underestimation or bias of the classifiers. 

We are now exploring how sensitive underestimation is to distribution variations in the class label and the sensitive feature. Our next step is to develop strategies to mitigate underestimation. 

\section*{Acknowledgements}
This work was funded by Science Foundation Ireland through the SFI Centre for Research Training in Machine Learning (Grant No. 18/CRT/6183)

\bibliographystyle{abbrv}  
\bibliography{bias}  
\end{document}